\documentclass[10pt,twocolumn,letterpaper]{article}

\usepackage{cvpr}              

\usepackage[accsupp]{axessibility}
\usepackage[dvipsnames]{xcolor}
\usepackage[utf8]{inputenc} 
\usepackage[T1]{fontenc}    
\usepackage{url}            
\usepackage{caption}
\usepackage{subcaption}
\usepackage{graphicx}
\usepackage{amsmath}
\usepackage{amssymb}
\usepackage{amsthm}
\usepackage{multirow,makecell}
\usepackage{booktabs}       
\usepackage{pgf, tikz}
\usepackage{bm}
\usepackage{dsfont}
\usepackage{mathtools}
\usepackage{float}
\usepackage{filecontents}
\usepackage{standalone}
\usepackage{algorithm}
\usepackage{algorithmic}
\usepackage{amsfonts}       
\usepackage{nicefrac}       
\usepackage{microtype}      
\usepackage{xcolor}         
\usepackage{wrapfig}

\DeclareMathOperator*{\argmax}{arg\,max}

\definecolor{cvprblue}{rgb}{0.21,0.49,0.74}
\usepackage[pagebackref,breaklinks,colorlinks,citecolor=cvprblue]{hyperref}

\def\paperID{5113} 
\def\confName{CVPR}
\def\confYear{2024}

\expandafter\def\expandafter\normalsize\expandafter{%
    \normalsize%
    \setlength\abovedisplayskip{4pt}%
    \setlength\belowdisplayskip{4pt}%
}

\title{Versatile Navigation under Partial Observability 
via Value-guided Diffusion Policy}

\author{\textbf{Gengyu Zhang}$^1$ \quad \textbf{Hao Tang}$^2$ \quad \textbf{Yan Yan}$^{1,}$\footnotemark[2]\\
$^1$Department of Computer Science, Illinois Institute of Technology, USA \\
$^2$Robotics Institute, Carnegie Mellon University, USA \\ 
{\tt\small gzhang32@hawk.iit.edu, bjdxtanghao@gmail.com, yyan34@iit.edu}
}

\begin{document}
\maketitle

\renewcommand{\thefootnote}{\fnsymbol{footnote}}
\footnotetext[2]{Corresponding author}

\begin{abstract}
Route planning for navigation under partial observability plays a crucial role in modern robotics and autonomous driving. Existing route planning approaches can be categorized into two main classes: traditional autoregressive and diffusion-based methods. The former often fails due to its myopic nature, while the latter either assumes full observability or struggles to adapt to unfamiliar scenarios, due to strong couplings with behavior cloning from experts. To address these deficiencies, we propose a versatile diffusion-based approach for both 2D and 3D route planning under partial observability. Specifically, our value-guided diffusion policy first generates plans to predict actions across various timesteps, providing ample foresight to the planning. It then employs a differentiable planner with state estimations to derive a value function, directing the agent's exploration and goal-seeking behaviors without seeking experts while explicitly addressing partial observability. During inference, our policy is further enhanced by a best-plan-selection strategy, substantially boosting the planning success rate. Moreover, we propose projecting point clouds, derived from RGB-D inputs, onto 2D grid-based bird-eye-view maps via semantic segmentation, generalizing to 3D environments. This simple yet effective adaption enables zero-shot transfer from 2D-trained policy to 3D, cutting across the laborious training for 3D policy, and thus certifying our versatility. Experimental results demonstrate our superior performance, particularly in navigating situations beyond expert demonstrations, surpassing state-of-the-art autoregressive and diffusion-based baselines for both 2D and 3D scenarios.
\end{abstract}
\vspace{-0.4cm}    
\section{Introduction}
\label{sec:intro}


Navigation is a critical component in mobile robotics and autonomous driving dependent on sequential planning, a process of evaluating and selecting an action sequence that most effectively achieves a specific goal. However, traditional autoregressive planning methods for navigation, as mentioned in~\cite{autoregr-tamar2016VIN,autoregr-karkus2017QMDP-net,autoregr-ishida2022calvin}, face two significant limitations. First, they select actions sequentially, where each decision is based on the previous one and the consequent state transitions.
\begin{figure}[!t]\small
    \centering
    \includegraphics[width=0.9\linewidth]{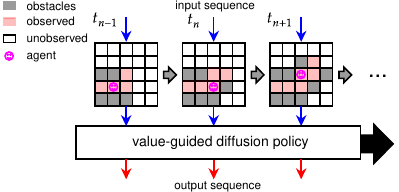}
    \caption{Our value-guided diffusion policy under partial observability. It processes local partial observations to generate action sequences adaptable for both 2D and 3D scenarios.}
    \label{fig:general_process}
    \vspace{-0.4cm}
\end{figure}
This step-by-step approach is not well-suited for tasks with longer horizons, as it lacks foresight. The problem worsens in partially observable settings, where increased uncertainty introduces greater computational demands to solve complex mathematical frameworks~\cite{pomdp-sondik1971optimal,pomdp-kaelbling1998planning}. Additionally, the necessity for instantaneous decision-making in real-time navigation can be at odds with the once-per-step inference rate of traditional planning methods. Second, these methods often require a substantial volume of data to learn effective policies for 3D navigation. In practice, however, gathering large datasets can be impractical due to environmental, logistical, or equipment constraints, resulting in a limited set of offline data. When faced with such data deficiency, traditional methods tend to yield suboptimal performance.

To overcome the limitations carried by autoregressive planning, we explore trajectory-level behavior synthesis. This novel approach capitalizes on the capabilities of generative models, particularly diffusion models~\cite{diffusion-janner2022diffuser,diffuison-pearce2023imitating,diffusion-liang2022adaptdiffuser,diffusion-chi2023diffusionpolicy,diffusion-ajay2023is,diffusion-margolis2022walk}. Unlike autoregressive methods that generate actions sequentially, diffusion-based approaches synthesize entire action trajectories simultaneously, enhancing multi-step planning efficiency during inference. However, to our best knowledge, no existing work of this class has actively explored their effectiveness under partial observability. Thus, through a significant modification, we adapt this concept for use in partially observable environments. We model the navigation with an approximated partially observable Markov decision process (POMDP). This involves embedding a state estimation module in training a differentiable planner, which learns a value function to guide the agent's policy planning.
This value function is derived by estimating the underlying decision model of expert demonstrations during training and iteratively computing optimal values during inference. \cref{fig:general_process} illustrates the plan generation module, in which the diffusion policy generates a plan for certain future timesteps in a closed-loop manner conditioned on the observed partial environment map. A plan in this context is essentially an action trajectory — a series of temporally sequential actions derived from a specific policy and state transition dynamics. The value function demonstrated in \cref{fig:reward-function} and \cref{fig:value-iteration} ensures that the generated plans lead to at least near-optimal outcomes through trajectory optimization.

To overcome the challenge of data scarcity in 3D realistic navigation scenes, we propose adapting inputs into a format amenable to models trained on 2D data, allowing us to apply policies learned from the 2D domain to navigate in 3D environments. This is predicated on the abundance of 2D data, which ensures that a robust policy for the 2D domain can be learned. By constructing a point cloud from first-person-view (FPV) RGB-D inputs and transforming it to meet 2D standards, we can preserve the performance of the 2D policy in the 3D navigation. This transformation involves semantic segmentation of the point cloud using pre-trained models~\cite{segment-yang2023swin3d,segment-deng2023pointvector}, followed by projecting it onto a bird-eye view (BEV) grid map based on the result of segmentation. Consequently, the high-dimensional RGB-D inputs are converted into grid maps, serving as the basis for our diffusion policy to generate action plans. The 2D policy, already trained, can then infer actions for 3D scenes.

We evaluate the efficacy of our method with two established frameworks: an autoregressive planner, CALVIN~\cite{autoregr-ishida2022calvin}, and a diffusion-based behavior cloner, Diffusion Policy~\cite{diffusion-chi2023diffusionpolicy}. Extensive experiments demonstrate the superiority of our method over these baselines in 2D mazes and real-world 3D indoor navigation scenes. Notably, the policy trained on 2D mazes is directly applicable to 3D settings by projecting the point cloud to 2D BEV plane, showing impressive scalability without additional training while still maintaining performance on par with the baselines. Further enhancements include training the model with both BEV grid maps converted from point cloud and egocentric RGB images, with supervision from a limited set of expert demonstrations in point cloud navigation. This dual-conditioning approach has been shown to boost the effectiveness of the policy. 
\section{Related Work and Preliminary}
\label{sec:related_work}
\noindent\textbf{Differentiable planning.} The concept of deep differentiable planning, a method that facilitates online plan generation and backpropagation of errors through these plans to train transition and reward estimators, was initially introduced by Value Iteration Networks (VINs) \cite{autoregr-tamar2016VIN,autoregr-deploy-schleich2019,autoregr-deploy-nie2021capability}. This approach is often employed in offline reinforcement learning~\cite{offlinerl-levine2020offline} and imitation learning where data is limited. Subsequent works adapted VINs to partially observable scenarios, assuming a complete environmental map for localization tasks \cite{autoregr-karkus2017QMDP-net}, and substituted the max pooling operation, which in VINs realizes the maximization in the Bellman equation of MDPs, with an LSTM structure. A recent enhancement to VINs introduced an additional mask for the explicit exclusion of invalid actions, thereby preventing collisions and allowing for more effective long-horizon navigation~\cite{autoregr-ishida2022calvin}.

\noindent\textbf{Diffusion models and diffusion policies.} Diffusion models, as a prominent class of generative models, formulate the data creation process as an iterative denoising procedure~\cite{diffusion-model-sohl2015deep,diffusion-model-ho2020denoising}. This approach can be interpreted as the parameterization of the gradients of data distribution~\cite{diffusion-model-song2019generative,diffusion-model-song2021scorebased,diffusion-model-Karras2022edm}, thereby linking diffusion models with score matching~\cite{diffusion-model-hyvarinen2005estimation} and Energy-Based Models (EBMs) \cite{diffusion-model-du2019implicit,diffusion-model-nijkamp2019learning}. The iterative, gradient-based sampling is particularly conducive to flexible conditioning \cite{diffusion-model-dhariwal2021diffusion,diffusion-model-nichol2021improved} and compositionality~\cite{diffusion-model-du2020compositional}. This led to the emergence of a promising new category of methods that harness the potential of diffusion models to extract effective behaviors from heterogeneous datasets and plan for unobserved scenarios during the training phase. Some of these approaches focus on the practical application of diffusion models for control policy behavioral cloning~\cite{diffusion-chi2023diffusionpolicy} or diffusion policy analysis in simulated environments~\cite{diffuison-pearce2023imitating}. Others explore the use of diffusion models in planning contexts, integrating a value function to facilitate planning for unseen tasks~\cite{diffusion-janner2022diffuser,diffusion-wang2023diffusion,diffusion-chen2023highfidelity}. Researchers also utilize diffusion models in robot learning in conjunction with physics-augmented simulations. This approach is instrumental in designing and developing varied and functional soft robot systems, with an emphasis on their morphology and control mechanisms~\cite{diffusion-wang2023diffusebot}.


Diffusion models posit data generation as an iterative denoising process, $p_{\theta}(\mathbf{x}_{i-1}|\mathbf{x}_i)$, reversing the forward diffusion process $q(\mathbf{x}_i|\mathbf{x}_{i-1})$ which consistently adds noise to the original data sample. This process is also known as Stochastic Langevin Dynamics~\cite{langevin-welling2011bayesian}. Here, $\mathbf{x}_i$ in the diffusion (forward) and reverse processes denotes the noisy data at the diffusion step $i$. The target data distribution that DPMs aim to recover from Gaussian noise, along with the corresponding denoising process, are as follows:
\begin{equation}
    \begin{aligned}
        p_{\theta}(\mathbf{x}_{0:N})&=p(\mathbf{x}_N)\prod_{i=1}^N p_{\theta}(\mathbf{x}_{i-1}|\mathbf{x}_i), \\
        p_{\theta}(\mathbf{x}_{i-1}|\mathbf{x}_i)&=\mathcal{N}(\mathbf{x}_{i-1};\bm{\mu}_{\theta}(\mathbf{x}_i,i),\mathbf{\Sigma}_{\theta}(\mathbf{x}_i,i)),
    \end{aligned}
\end{equation}
where $\mathcal{N}$ signifies a Gaussian distribution with mean $\bm{\mu}_{\theta}(\mathbf{x}_i,i)$ and variance $\mathbf{\Sigma}_{\theta}(\mathbf{x}_i,i))$, each is a function of the data sample $\mathbf{x}_i$ and step $i$. We adopt the notation used in~\cite{diffusion-janner2022diffuser} that denotes the number of diffusion steps with $N$ and each step with $i$, distinguishing from $T$ and $t$ used for planning timesteps.
The iterative sampling process of diffusion models facilitates flexible conditioning, allowing auxiliary guides to adjust the sampling procedure to retrieve trajectories with high returns or satisfy specific constraints. Incorporating trajectory optimization into the modeling process permits diffusion policies to enhance the performance of learned models in decision-making tasks.

\noindent\textbf{Planning under partial observability.} POMDP is a widely recognized mathematical framework for modeling decision-making scenarios with imperfect observation. In such contexts, an agent lacks direct access to the complete information necessary to fully describe the state of the system. A POMDP is formally defined as the tuple ($\mathcal{S},\mathcal{A},\Omega,\mathcal{T},\mathcal{R},\mathcal{O},\gamma$), where $\mathcal{S}$, $\mathcal{A}$, and $\Omega$ are the discrete state, action, and observation spaces, respectively. The state-transition function, $\mathcal{T}:\mathcal{S}\times\mathcal{A}\times\mathcal{S}\rightarrow[0,1]$, denotes the probability of transitioning from state $s$ to $s^{\prime}$. The reward function, $\mathcal{R}:\mathcal{S}\times\mathcal{A}\rightarrow\mathbb{R}$, quantifies the immediate reward gained by executing action $a$ in state $s$. The observation function, $\mathcal{O}:\mathcal{S}\times\mathcal{A}\times\Omega\rightarrow[0,1]$, specifies the likelihood of receiving observation $o$ in state $s$ by taking action $a$. Lastly, $\gamma$ is the discount factor used in the Bellman equation for iterative optimal value derivation.


Under partial observability, since the agent cannot directly observe the underlying physical state, it instead maintains a probability distribution over $\mathcal{S}$, namely the belief state, that indicates its confidence about which state it is in. Over time, the belief state is updated in a Bayesian manner following Eq.~\eqref{eq:pomdp-belief-update}, where it is first updated by action $a$ and transition dynamics $\mathcal{T}$, and then corrected by observation $o^{\prime}$ and observation function $\mathcal{O}$. $\eta$ is the normalization factor. This procedure is also called the state estimation.
\begin{equation}
    \label{eq:pomdp-belief-update}
    b(s^{\prime})=\eta\mathcal{O}(s^{\prime},a,o^{\prime})\sum\nolimits_{s\in\mathcal{S}}\mathcal{T}(s^{\prime}|s,a)b(s),
\end{equation}
However, this presents a significant computational challenge for the optimal policy derivation of POMDPs. Consider a system with $n$ physical states; the policy $\pi$ must be defined across a $(n-1)$-dimensional continuous belief space, making it prohibitively expensive to solve by standard value or policy iteration. This challenge, known as the \emph{curse of dimensionality}, is one of the two primary factors that contribute to the computational intractability of solving POMDPs exactly~\cite{qmdp-2}. The other factor, termed the \emph{curse of history}, arises from the exponential growth in the number of distinct action-observation histories to be evaluated for policy optimization as the planning horizon extends.

To mitigate these challenges, we adopt QMDP~\cite{qmdp-1,qmdp-2}, a heuristic that offers an approximate solution to POMDPs, effectively addressing both the curse of dimensionality and the curse of history. QMDP employs a simplified model that considers partial observability at the current planning step but assumes full observations for subsequent steps, which reduces computational complexity while still accounting for the uncertainty, thus offering a computationally efficient, approximate solution scaling to larger problems.

QMDP obtains the optimal $Q$ function by solving the corresponding fully observable MDP via iterating the following Bellman equation until convergence.
\begin{equation}
    \label{eq:QMDP_bellman}
    \mkern-28muQ^{k+1}(s,a)=\mathcal{R}(s,a)+\gamma\sum\nolimits_{s^{\prime}}\mathcal{T}(s^{\prime}|s,a)\max_{a^{\prime}}Q^k(s^{\prime},a^{\prime}),\!\!
\end{equation}
where $k\in[1,K]$ denotes the current iteration round. Finally, we obtain the QMDP policy by:
\begin{equation}
    \label{eq:QMDP_policy}
    \pi(b)=\argmax_a\sum\nolimits_{s}Q^K(s,a)b(s).
\end{equation}
\vspace{-0.8cm}
\section{Methodology}
\label{sec:method}

This section introduces our novel navigation framework, which harnesses diffusion models for generating action trajectories in complex, partially observable environments. This framework comprises 1) the diffusion policy module and 2) the value network. The diffusion policy module, outlined in \cref{subsec:method:diffusion_policy}, lies in generating plans based on partial environment maps, enhancing the agent's decision-making as it gathers more environmental data. Our closed-loop planning process, underpinned by receding horizon control, ensures a smooth and coherent action trajectory formulation. To address the limitations of behavioral cloning in dynamic settings, we incorporate the value guidance as detailed in \cref{subsec:method:value_guidance}. This enhancement, critical in complex environments, drives the agent away from obstacles and dead ends. Our method integrates state estimation with QMDP to approximate the optimal value function and reinforce the policy's efficacy. We train these two modules separately and incorporate them for inference. A unique aspect of our approach, described in \cref{subsec:method:3d_to_2d_projection}, involves adapting our robust 2D policy for 3D environments. We transform 3D RGB-D inputs into 2D BEV maps, allowing for a seamless transfer of 2D navigation policy to 3D scenarios. This method overcomes the challenges posed by the deficiency of real-world 3D data, thus facilitating efficient and accurate navigation in various settings.
\begin{figure}[!t]\small
    \centering
    \includegraphics[width=0.85\linewidth]{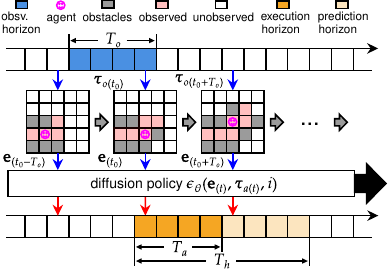}
    \caption{The architecture of diffusion-model-based plan generator. The top sequence represents local observations over time. The grids in the middle form the sequence of cumulative partial maps, which sufficiently encapsulate the agent's long-term memory and environment features. The bottom sequence represents the generated plan in the form of action trajectories. During training, the input of the framework at timestep $t$ consists of the partial map, $\bm{e}_{(t)}$, and expert action trajectory, ${\bm{\tau}_a}_{(t)}$; during inference, the input comprises $\bm{e}_{(t)}$ and a Gaussian noise of the same shape as ${\bm{\tau}_a}_{(t)}$.}
    \label{fig:arch_diffusion_policy}
    \vspace{-0.6cm}
\end{figure}

\subsection{Problem Formulation}
\label{subsec:mathod:problem_setting}
We aim to address a trajectory optimization problem similar to that in~\cite{diffusion-janner2022diffuser} but under partial observability. In a discrete-time control system, where the dynamics are defined as $s_{t+1}=f(s_t,a_t)$ at state $s_t$ and given action $a_t$, we seek to search for a plan, in the form of an action sequence, $\bm{a}_{0:T}^{\ast}$, that maximizes an objective function $\mathcal{J}$. This objective function is factorized over per-timestep $Q$ values, $Q_t(b_t,a_t)$:
\begin{equation}
    \begin{aligned}
        \mathcal{J}(b_0,\bm{a}_{0:T})&=\sum_{t=0}^TQ_t(b_t,a_t)=\sum_{t=0}^TQ_t(s_t,a_t)b_t(s_t),\\
    \bm{a}_{0:T}^{\ast}&=\argmax_{\bm{a}_{0:T}}\mathcal{J}(b_0,\bm{a}_{0:T}).
    \end{aligned}
\end{equation}
In our model, the belief $b_t$ is updated according to \cref{eq:pomdp-belief-update} throughout the planning horizon $T$. We define the action trajectory at time $t$ as $\bm{\tau}_{a,(t)}=(a_t,a_{t+1},\ldots,a_{t+T-1})$, which the diffusion model generates, conditioned on the partially observed environment $\bm{e}_{(t)}$. This environment map, $\bm{e}_{(t)}$, compiles the trajectory of local observations from timestep 0 to $t$. Given a previously obtained map $\bm{e}_{(t-T_o)}$, we have $\bm{e}_{(t)}=(\bm{e}_{(t-T_o)}, \bm{\tau}_{o,(t)})$, where $\bm{\tau}_{o,(t)}=(o_{t-T_o+1}, o_{t-T_o+2}, \ldots, o_t)$ and $T_o$ represent the observation horizon, as illustrated in \cref{fig:arch_diffusion_policy}. In our framework, $T_o$ aligns with $T_a$, the horizon for action execution, which is further detailed in \cref{subsec:method:diffusion_policy}.

\subsection{Diffusion-model-based Plan Generation}
\label{subsec:method:diffusion_policy}

As depicted in \cref{fig:arch_diffusion_policy}, our framework utilizes a diffusion model to generate action trajectories from timestep $t$. These trajectories are conditioned on the partial environment map, $\bm{e}_{(t)}$, which aggregates information observed up to $t$. As a result, as the agent continues to explore its surroundings, its understanding of the overall environment gradually enhances, facilitating more rational decision-making.

The process of plan generation in our framework operates in a closed-loop manner. In each iteration, we input the partial environment map into the diffusion model. This map, which encapsulates sufficient statistics of the observation history, acts as the key condition, steering the diffusion model's conditional generation process. In a partially observable scenario, the agent uncovers new areas incrementally, gradually removing the mist and enriching the existing map with additional world information. To encourage the temporal coherence and smoothness of formulating action trajectories during planning, we adopt the receding horizon control strategy~\cite{mayne1988receding}. In practice, at each timestep $t$, the policy processes the current partial environment map, $\bm{e}_t$, forecasts actions for the next $T_h$ steps (the prediction horizon), and then implements the initial $T_a$ steps (the execution horizon) before the next planning cycle, thereby streamlining the decision-making process.

We train a diffusion model to learn a robust policy that captures the conditional distribution $p(\bm{\tau}_{a,(t)}|\bm{e}_{(t)})$. We formalize the denoising diffusion process, theoretically in the form of $\bm{\tau}^{i-1}_{a,(t)}\sim p_{\theta}(\bm{\tau}^{i-1}_{a,(t)}|\bm{\tau}^i_{a,(t)})$ as
\begin{equation}
    \label{eq:ddpm_policy_denoising}
    \bm{\tau}^{i-1}_{a,(t)}=\delta(\bm{\tau}^i_{a,(t)}-\alpha\bm{\epsilon}_{\theta}(\bm{e}_{(t)},\bm{\tau}^i_{a,(t)},i)+\bm{\epsilon}^i),\text{where~}\bm{\epsilon}^i\sim\mathcal{N}(\mathbf{0},\mathbf{I})
\end{equation}
Here, $\delta$ denotes the step size, $\alpha$ represents the learning rate, and $i$ is the diffusion step. We employ a mean squared error to account for the loss function:
\begin{equation}
    \label{eq:loss_ddpm_policy}
    \mathcal{L}_{\rm{MSE}}(\theta)=\mathbb{E}_{i,\bm{\epsilon}}[\lVert\bm{\epsilon}^i-\bm{\epsilon}_{\theta}(\bm{e}_{(t)}, \bm{\tau}^0_{a,(t)}+\bm{\epsilon}^i,i)\rVert^2].
\end{equation}

\subsection{Value-guided Exploration-safe Planning}
\label{subsec:method:value_guidance}
Employing only the diffusion plan generator is essentially behavioral cloning~\cite{diffusion-chi2023diffusionpolicy,diffuison-pearce2023imitating} under partial observability. This adaptation, however, retains inherent weaknesses in navigating complex and dynamic environments. A notable challenge arises when a policy conditioned on limited environmental data inadvertently leads the agent to a dead end. Since expert demonstrations do not cover such circumstances, the agent might struggle to backtrack and seek alternate paths. This exposes a common downside to diffusion-model-based behavioral cloning methods: a lack of deep environmental understanding.

Incorporating value guidance to direct the agent to the goal while avoiding obstacles presents an effective solution to this challenge. Several studies~\cite{diffusion-janner2022diffuser,diffusion-wang2023diffusion,diffusion-chen2023highfidelity} have explored implementing value guidance in fully observable settings to enhance diffusion policies. To tackle partial observability, we augment one of our baselines~\cite{autoregr-ishida2022calvin} by integrating a state estimation module, utilizing QMDP to approximate the optimal value function under partially observable conditions.

\begin{figure}[!tbp]\small
    \centering
    \includegraphics[width=0.9\linewidth]{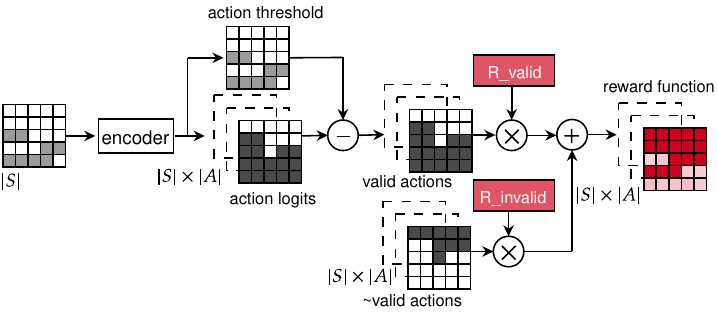}
    \caption{Reward function conditioned on the partial environmental map. The model learns a valid action mask that filters out invalid actions using soft thresholding. This learned embedding is subsequently used to construct the reward function.}
    \label{fig:reward-function}
    \vspace{-0.4cm}
\end{figure}
\noindent\textbf{Value function with state estimation.} The state estimation module implements a Bayesian filter that maps a belief, an action, and an observation to the subsequent belief according to \cref{eq:pomdp-belief-update}. This module comprises two components: 1) a prediction module that learns a state transition function that predicts the next belief in a translation-invariant manner for each action, and 2) a belief correction module that weights the updated belief by a jointly learned observation function.
\begin{figure}[!tbp]\small
    \centering
    \includegraphics[width=0.9\linewidth]{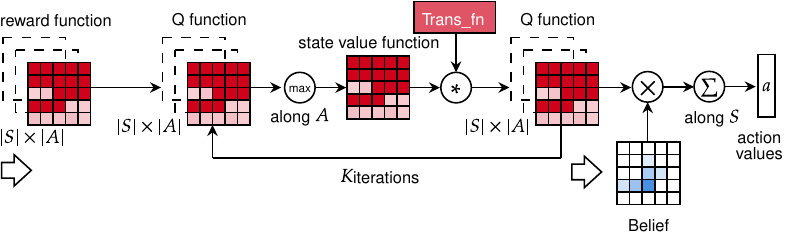}
    \caption{QMDP value iteration module. The learned reward function undergoes $K$ rounds of iterations, consisting of alternating maximization over actions and convolution with the transition function $\hat{T}_m$. The outcome, soft-indexed by the current belief, derives the final action values of this QMDP planner.}
    \label{fig:value-iteration}
\end{figure}
For the value function module (\cref{fig:reward-function}), we first train an action validity estimator to explicitly recognize valid and invalid actions for each state. To achieve this, our framework learns the logarithmic probability of each action, $\hat{A}_{\rm{logit}}(s,a)$, and an action threshold function, $\hat{A}_{\rm{thresh}}(s)$, both conditioned on the partial environment map. Specifically, $\hat{A}_{\rm{logit}}$ and $\hat{A}_{\rm{thresh}}$ are part of the same learned embedding, comprising $|A|+1$ channels, where $|A|$ denotes the size of the action space. The first $|A|$ channels form $\hat{A}_{\rm{logit}}$, while the last channel serves as $\hat{A}_{\rm{thresh}}$. Thus, we derive the valid action, $\hat{A}$, by applying a soft-threshold to $\hat{A}_{\rm{logit}}(s,a)$ using $\hat{A}_{\rm{thresh}}(s)$:
\begin{equation*}
    \hat{A}(s,a)=\sigma(\hat{A}_{\rm logit}(s,a)-\hat{A}_{\rm thresh}(s)),
\end{equation*}
where $\sigma$ is the Sigmoid function. Once we acquire $\hat{A}$ and its negation, $\sim\!\!\hat{A}$, we construct the final reward function by merging them with the separately learned reward parameters: $\hat{R}_m$ for valid actions and $\hat{R}_f$ for invalid ones. In this reward learning framework, $\hat{R}_f$ is typically assigned large negative values, effectively reducing collisions during navigation. We formally define our new reward function as follows:
\begin{equation}
    \label{eq:proposed_reward_function}
    \begin{aligned}
        \mathcal{R}(s,a)&=\hat{R}_f(1-\hat{A}(s,a)) + \\
        &\hat{A}(s,a)\sum\nolimits_{s^{\prime}}\hat{T}_m(s^{\prime}|s,a)\hat{R}_m(s,a,s^{\prime}),
    \end{aligned}
\end{equation}
where $\hat{T}_m$ estimates the subsection of true state transition pertaining to valid actions. Given this enhanced reward function, the subsequent value iteration module, used to compute the optimal value function, adopts the design as depicted in \cref{fig:value-iteration}. After $K$ iterations, the resulting $Q$ function is soft indexed by current belief to derive the approximated QMDP optimal value function, which we use to guide the diffusion policy in inference.
\setlength{\textfloatsep}{5pt}

\begin{algorithm}[!t] \small
    \captionsetup{font=small}
    \caption{Best Plan Candidate Backtracking}
    \label{algo:candidate_backtracking}
    \begin{algorithmic}[1]
        \REQUIRE value function $\mathcal{Q}_{(t)}(s,a;\theta)$, diffusion policy $\bm{\epsilon}_{\theta}$, best plan memory $\mathcal{C}$, best plan $\hat{\bm{\tau}}_{a,(t)}^{\ast}$
        \STATE $\hat{\bm{\tau}}_{a,(t^{\prime})}^{\ast}\gets\text{empty}$,~$\mathcal{C}\gets\emptyset$
        \FOR{$t = 0, 1, \ldots, T$}
            \STATE $\mathcal{C}\gets\bm{\epsilon}_{\theta}(\bm{e}_{(t)},\bm{\tau}^N_{a,(t)},N)$ executed for $L$ times
            \IF{$\hat{\bm{\tau}}_{a,(t^{\prime})}$ is not empty}
                \STATE $\mathcal{C}\gets\mathcal{C}\cup\{\hat{\bm{\tau}}_{a,(t^{\prime})}^{\ast}\}$
            \ENDIF
            \STATE $\hat{\bm{\tau}}_{a,(t)}^{\ast}\gets\argmax_{\bm{\tau}\in\mathcal{C}}\frac{1}{|\tau|}\sum_{i=0}^{|\tau|-1}Q_{(t)}(s_{\bm{\tau}_i},a_{\bm{\tau}_i};\theta)b_{(t+i)}(s_{\bm{\tau}_i})$  \COMMENT{\cref{eq:value-guided-plan-selection}}
            \STATE $\Delta t\gets\min(T_a,|\hat{\bm{\tau}}_{a,(t)}^{\ast}|)$
            \STATE Execute first $\Delta t$ actions of $\hat{\bm{\tau}}_{a,(t)}^{\ast}$
            \STATE Remove first $\Delta t$ actions from $\hat{\bm{\tau}}_{a,(t)}^{\ast}$
            \STATE $\hat{\bm{\tau}}_{a,(t^{\prime})}\gets\hat{\bm{\tau}}_{a,(t)}^{\ast}$
        \ENDFOR
    \end{algorithmic}
\end{algorithm}

\noindent\textbf{Value-guided plan selection.} We use a well-learned value function to guide the diffusion policy. The stochastic nature of diffusion models, stemming from noise sampling, enables us to generate diverse plans given the same conditions. By calculating the sum of action values along each plan, we determine the values of a set of action sequences and select the one with the highest value to execute. This design substantially enhances the navigation's success rate. 

However, while the receding horizon control (\cref{subsec:method:diffusion_policy}) used in the diffusion plan generation encourages temporal coherence of predicted multi-step plans and strengthens their robustness against latency, it can lead to suboptimal plans. Specifically, when the policy predicts $T_h$ steps of actions and executes the first $T_a$ steps, a left-behind but optimal action $a$ might be overwritten by some suboptimal $a^{\prime}$ in the re-planning starting from the end of the execution sequence. This issue occurs due to a covariate shift of testing observations from expert demonstration and the diffusion process's stochasticity. To cope with this issue, we propose maintaining a buffer to backtrack the best action trajectory candidates from the past, preserving optimal actions in at least one candidate to avoid inevitable failure. \cref{eq:value-guided-plan-selection} demonstrates the criterion of selecting the optimal plan at timestep $t$. In this equation, $\hat{\bm{\tau}}_{a,(t)}^{\ast}$ represents the predicted optimal action trajectory selected from a set of trajectories $\mathcal{C}$, $\hat{\bm{\tau}}_{a,(t^{\prime})}^{\ast}$ is the best plan candidate selected last time with executed actions removed, where $t^{\prime}$ is the last timestep a plan is selected. $Q_{(t)}$ refers to the learned $Q$ function at the current timestep. The pseudocode for the backtracking process is provided in \cref{algo:candidate_backtracking}, where $N$ is the total number of diffusion steps for plan generation, and $L$ is the number of candidates to generate each time. Note that we only apply backtracking during inference. Hence, the refined policy becomes:
\begin{equation}
    \label{eq:value-guided-plan-selection}
     \mkern-36mu\hat{\bm{\tau}}_{a,(t)}^{\ast}=\argmax_{\bm{\tau}\in\mathcal{C}\cup\{\hat{\bm{\tau}}_{a,(t^{\prime})}^{\ast}\}}\frac{1}{\bm{|\tau|}}\sum_{i=0}^{|\bm{\tau|}-1}Q_{(t)}(s_{\bm{\tau}_i},a_{\bm{\tau}_i};\theta)b_{(t+i)}(s_{\bm{\tau}_i}).\!\!\!
\end{equation}

\vspace{-0.2cm}

\subsection{2D to 3D Policy Transfer}
\label{subsec:method:3d_to_2d_projection}
\begin{figure}[!tbp]
    \centering
    \includegraphics[width=0.95\linewidth]{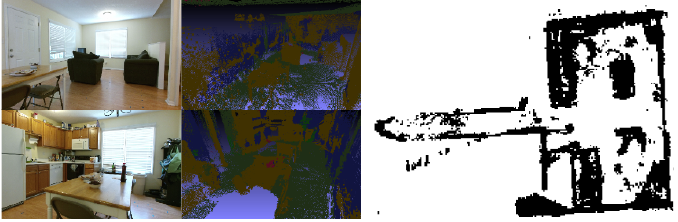}
    \caption{An illustration of constructing a point cloud for a given scene and its subsequent projection onto a BEV map. In this specific example, objects such as the table, chair, and various other furniture pieces in the kitchen, the two sofas and television cabinets in the living room, and the surrounding walls are identified as obstacles on the BEV map. Conversely, areas of the floor that remain uncovered by any objects are designated as free space.}
    \label{fig:pc_to_bev}
\end{figure}

3D data scarcity poses a significant challenge due to constraints in the real world. Training models on such sparse datasets often leads to overfitting, compromising the ability to generalize to new environments. To circumvent this, we leverage the robust policy developed for the 2D domain, aiming for a zero-shot application to 3D environments. This necessitates transforming 3D inputs into a format compatible with our established 2D policy. In the context of 3D embodied navigation, the agent processes first-person-view RGB-D images. To convert these into 2D BEV maps, we first construct a point cloud from accumulated RGB-D data and then apply pre-trained semantic segmentation models~\cite{segment-yang2023swin3d} to categorize various elements (\cref{fig:pc_to_bev}). Key labels for crafting the 2D grid map include floors, indicating traversable areas, and walls or furniture, representing obstacles.

We follow a multi-step process to project these segmented components onto a BEV plane. Initially, we trim the point cloud along the $Z$-axis, which is absent in the BEV representation, to a fixed proportion and perform downsampling. Subsequently, we project the points onto a grid. Whether each cell is classified as free space or an obstacle hinges on the prevalence of points identified as the floor within it. This method allows us to replicate a 2D grid map analogous to those used in our 2D maze experiments, effectively bridging the gap between route planning of 2D and 3D navigation.
\section{Experiments}
\label{sec:experiments}


\subsection{Task Setups} 
\label{subsec:experiments:task_setting}
\noindent\textbf{GridMaze2D.} This classic domain provides diverse synthetic environments and tasks for evaluating our method. In this domain, the agent is expected to explore an arbitrary partially observable maze, find the goal, and execute a termination action, \emph{Done}, to finish the current task. If the agent terminates the task at the end of the goal, it successfully completes the mission. Otherwise, running into an obstacle, executing \emph{Done} in the wrong state, or failing to terminate the task all lead to failure. Each environment of this domain corresponds to a unique 2D maze map that presents a BEV of that environment. The observed partial maps are part of the maze map. Please see the Appendix for more details about the composition and generation of partial maps.

The valid action space contains eight directional movements and a termination action, all of which are categorical. Hence, we use the bit encoding technique~\cite{chen2022analog} to convert them into bit arrays for easy retrieval from Gaussian noise. During inference, we decode the sampled action trajectory back into categorical form. Please refer to the Appendix for technical details. The state space consists of the maze's full ($X,Y$) coordinates.

We simulate navigation in randomly generated mazes to collect expert trajectories. During each simulation, we record the agent's actions, positional coordinates (physical states), and partial environment maps (observation history) at each timestep throughout the trajectory. The expert, equipped with prior knowledge of the goal location, employs an informed search strategy like $A^{\ast}$ to navigate toward the goal. Upon gathering sufficient expert trajectories, we partition them into training and validation sets, ensuring that the environments in the validation data remain unseen while training.

\noindent\textbf{Active Vision Dataset.} This dataset for 3D embodied navigation enables interactive navigation using real image streams, as opposed to synthetic rendering. AVD consists of 19 indoor environments, densely captured by a robot navigating on a 30cm grid with 30$^\circ$ rotational increments. The comprehensive image set from each scene allows for the simulation of various trajectories with a certain degree of spatial granularity. Additionally, AVD provides bounding box annotations for object instances, a feature we utilize to assess semantic navigation performance. In this domain, the agent's objective is to navigate an indoor environment to locate and reach a specified object.

The action setup is similar to that in the GridMaze2D domain, where the agent has the option to move in any of eight directions. Upon identifying and reaching the target object, the agent must actively execute \emph{Done} command to terminate the current task. This time, we define the state space as the cell coordinates of the BEV map corresponding to each scene. A key difference from the GridMaze2D setup is that actions that lead to collisions with obstacles do not cause instant failure. Instead, the agent remains at the point until it navigates a clear path.

To assess the effectiveness of zero-shot policy transfer from GridMaze2D to the AVD domain, facilitated by point cloud projection, we chose 8 out of the 19 scenes containing a Coca-Cola glass bottle as our validation set. To evaluate CALVIN and our retrained policy with additional RGB inputs, we adopt cross-validation, using one scene for validation and the others for training.


\subsection{Result Analysis}
For the GridMaze2D domain, we train our model on 15$\times$15 mazes with view range 2. To evaluate robustness against different observability levels, we test the learned policy across three view range settings.
\begin{table}[!tbp]
    \centering
    \small
    \setlength{\tabcolsep}{0.8pt}
    \scalebox{0.7}{
    \begin{tabular}{lccc}
    \toprule
    & CALVIN \cite{autoregr-ishida2022calvin} & Diffusion Policy \cite{diffusion-chi2023diffusionpolicy} & Ours \\
    \midrule
    15$\times$15 (vr=1) & 0.832$\pm$0.030 & 0.024$\pm$0.015 & \textbf{0.886$\pm$0.011} \\
    15$\times$15 (vr=2) & 0.855$\pm$0.030 & 0.060$\pm$0.022 & \textbf{0.906$\pm$0.010} \\
    15$\times$15 (vr=3) & 0.900$\pm$0.026 & 0.110$\pm$0.031 & \textbf{0.911$\pm$0.013} \\
    \midrule
    20$\times$20 (vr=2) & 0.658$\pm$0.016 & 0.012$\pm$0.010 & \textbf{0.713$\pm$0.020} \\
    30$\times$30 (vr=2) & 0.326$\pm$0.030 & 0.000$\pm$0.000 & \textbf{0.624$\pm$0.032} \\
    \bottomrule
    \end{tabular}}
    \caption{For each method, we train the model on $15\times15$ mazes with a view range equal to $2$ and evaluate in three different maze sizes with three different view range settings. The results demonstrate their scalability to unseen and larger environments. Overall, our approach has better performance.}
    \label{table:tab_results_grid_maze}
\end{table}
To evaluate the generalization capability, we test our model across unseen 15$\times$15, 20$\times$20, and 30$\times$30 mazes. We compare our approach regarding the success rate of completing the navigation task with two baseline methods: 1) CALVIN~\cite{autoregr-ishida2022calvin}, an autoregressive differentiable planner, and 2) Diffusion Policy~\cite{diffusion-chi2023diffusionpolicy}, a diffusion-based behavioral cloner. The results represent the mean and standard deviation across five trials, each encompassing 500 distinct maze simulations. For the AVD, we redeploy a model trained on 30$\times$30 mazes and then transform the input RGB-D images into a point cloud. It is then projected onto a 2D partial environment map in each planning step. To evaluate the pre-trained model's zero-shot transfer to real-world scenes, we only feed it with the partial map, the same as in 2D mazes. To assess the model retrained with RGB-D inputs, we first feed the FPV images into the additional feature extraction module and then concatenate the output embedding with the partial map as the final input to the model. In this setup, we compare our method with CALVIN and its variant that employs our 2D-to-3D policy transfer technique, deriving the mean and standard deviation of 5 trials, each comprising 50 simulations per scene.

We first analyze the overall performance of each methodology regarding success rate in different domains. The numerical results shown in \cref{table:tab_results_grid_maze} reveal that CALVIN achieves solid performance with a mean success rate of 0.855 on 15$\times$15 mazes with view range 2 (standard setup), implying that it learns a proficient value function and identifies a near-optimal policy based on it. However, when scaling to larger mazes, the performance of both two variants noticeably declines. The method achieves the best performance in only one test scene. This is likely due to the increased planning horizon. In more expansive environments, the planning horizon extends, requiring more rounds of value iteration to adapt effectively. Nevertheless, since the function approximation deprives the value iteration of its monotonic improvement property, simply applying the model inference for additional iterations does not always work in larger environments. CALVIN's performance in embodied indoor navigation (\cref{table:tab_avd_res}) is restricted by the small size of the dataset. The policy learned in scenes belonging to the training set is hard to generalize to unseen scenes in the validation set.

\begin{table}[!tbp]
    \small
    \centering
    \setlength{\tabcolsep}{2.0pt}
    \scalebox{0.8}{\begin{tabular}{lcccc}
    \toprule
    Scene                   & CALVIN-2D & CALVIN-3D   & Ours (Zero-shot)    & Ours (Retrain)\\
    \midrule
    Home\_001\_1         & 0.692$\pm$0.037      & 0.720$\pm$0.052  &        0.769$\pm$0.038   & \textbf{0.776$\pm$0.028}\\
    Home\_001\_2         & 0.627$\pm$0.037      & 0.640$\pm$0.048  &  0.655$\pm$0.033   & \textbf{0.732$\pm$0.030} \\
    Home\_002\_1         & 0.735$\pm$0.035      & 0.740$\pm$0.048  &  0.728$\pm$0.034   & \textbf{0.755$\pm$0.027} \\
    Home\_003\_1         & 0.606$\pm$0.042      & 0.642$\pm$0.060  &  0.638$\pm$0.041   & \textbf{0.686$\pm$0.031} \\
    Home\_003\_2         & 0.558$\pm$0.033      & 0.590$\pm$0.043  &  0.603$\pm$0.033   &          \textbf{0.622$\pm$0.030} \\
    Home\_004\_1         & 0.647$\pm$0.040      & 0.680$\pm$0.050  &  0.684$\pm$0.042   &          \textbf{0.695$\pm$0.036} \\
    Home\_007\_1         & 0.587$\pm$0.038      & \textbf{0.610$\pm$0.045}  & 0.584$\pm$0.039   & 0.601$\pm$0.035 \\
    Home\_010\_1         & 0.728$\pm$0.033      & 0.736$\pm$0.043  &  0.769$\pm$0.032   & \textbf{0.781$\pm$0.028} \\
    Mean succ. rate      & 0.635$\pm$0.032 & 0.682$\pm$0.047  & 0.679$\pm$0.040 & \textbf{0.706$\pm$0.032} \\
    \bottomrule
    \end{tabular}}
    \caption{Performance of CALVIN and our method in AVD's embodied navigation and object searching tasks, where the goal is to locate a Coca-Cola glass bottle in an indoor scene. It presents the agent's success rates across various scenes. Our method, which achieves comparable results to CALVIN in zero-shot policy transfer from the 2D domain, surpasses CALVIN in scenarios retrained with additional RGB inputs, with an exception in one scene.}
    \label{table:tab_avd_res}
\end{table}

Diffusion Policy, equivalent to our framework without value guidance, attains a far lower success rate in GridMaze2D. This behavioral cloning approach hinges solely on conditional diffusion for policy derivation, neglecting the value function's role. As the maze expands, diffusion policy's effectiveness further diminishes, failing all navigation tasks in 30$\times$30 mazes. This trend underscores the significance of value guidance in partially observable navigation, particularly when the target's location is unknown beforehand. Given the inferior performance, we exclude the diffusion policy from the comparative analysis in the more intricate AVD domain.

Our approach demonstrates a strong success rate of 0.906 in the standard setup of GridMaze2D, outperforming CALVIN and setting new state-of-the-art performance. Despite a performance dip in larger environments, the decline is gradual, underscoring our work's superior scalability. This success is largely due to the incorporation of multi-step action values in our value-guided plan selection for trajectory optimization (\cref{eq:value-guided-plan-selection}) instead of focusing solely on the next step. This approach effectively mitigates potential collisions or repetitions during navigation. In the AVD domain, the superiority of our approach becomes more evident. Independent of limited scenes for policy learning, our policy transferred from GridMaze2D backed by extensive training data demonstrates improved generality and robustness, leading to better performance as shown in \cref{table:tab_avd_res}.

\begin{figure}[!tbp]\small
    \centering
    \includegraphics[width=0.82\linewidth]{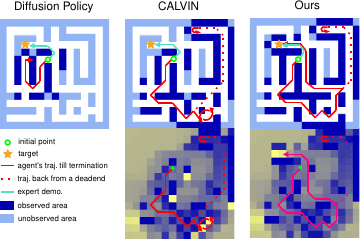}
    \caption{A scenario where three methods navigate the same maze. Diffusion Policy \textcolor{red}{(left)} collides with an obstacle after encountering a dead end, while CALVIN \textcolor{red}{(middle)} gets trapped in a repeating loop at a corner. Our approach \textcolor{red}{(right)}, however, successfully backtracks from a dead end and identifies an alternate path to the goal, demonstrating its superior performance. Please note that the expert demonstration is gathered under \textbf{full} observability,  with prior knowledge of the target's location. The heat maps illustrate the value learned at each spatial location by CALVIN and our framework, respectively, with brighter colors indicating higher values.}
    \label{fig:experiment_example}
\end{figure}

\cref{fig:experiment_example} illustrates a scenario where the three methods navigate the same maze. CALVIN falls into an indefinite loop due to the opposite actions suggested by the learned policy for observations of two consecutive steps. This is due to a combination of suboptimal modeling of the decision process and autoregressive single-step planning. Diffusion Policy fails early, especially after reaching a dead end, mainly due to two aspects. First, since the expert has full observation of the environment and is optimal, its demonstration for training never involves situations of encountering dead ends. Second, the behavior cloning essence of the Diffusion Policy is known to be effective in goal-conditioned planning. However, the agent cannot access the goal under partial observability until it is detected, significantly dropping the method's performance. On the contrary, our approach avoids loops by multi-step action prediction. It circumvents obstacles and safely backtracks from dead ends via effectively learned value guidance. This helps generalize the policy to unseen situations. Using the value as a guide also eliminates the need for access to the goal.

We then evaluate the robustness of the presented methods for different observability levels. As shown in \cref{table:tab_results_grid_maze}, a smaller view range, indicating a lower observability level, generally leads to performance degradation. Among the three methods, ours exhibits superior performance and adaptability. As the view range increases from vr=1 to vr=3, our method consistently outperforms the others and is the least sensitive to the variation. CALVIN also shows good robustness and scalability, though it does not match our approach's performance. The Diffusion Policy struggles significantly in comparison, showing the least robustness and lowest performance across all observabilities. This analysis underscores the effectiveness and reliability of our method.

\subsection{Ablation Study}
We conduct a series of ablation studies to assess the contribution of each core component of our framework to performance in 15$\times$15 grid mazes, and AVD embodied navigation. The full version can be represented as multi-samp.$+$val. guidance$+$best-plan memo.

\noindent\textbf{Effect of multi-sampling.} The Diffusion Policy~\cite{diffusion-chi2023diffusionpolicy}, employs single-sampling. In contrast, our approach samples multiple times for each planning. The inherent stochasticity of diffusion sampling generates varied outputs, from which we choose the most frequent outcome for execution using a voting mechanism. This strategy modestly increases the success rate, underscoring the advantages of leveraging diffusion models for multiple rounds of sampling.

\noindent\textbf{Effect of value guidance.} Leveraging multi-sampling,
\begin{table}[!t]
  \small
  \centering
  \scalebox{0.83}{\begin{tabular}{lcc}
    \toprule
    Ablation                                                   & GridMaze2D & AVD \\
    \midrule
    Full version                                          & \textbf{0.906$\pm$0.010} & \textbf{0.776$\pm$0.028}\\
    \midrule
    single-samp.                              &
    0.060$\pm$0.022 & 0.024$\pm$0.012\\
    multi-samp.$+$voting                        &
    0.114$\pm$0.025 & 0.082$\pm$0.026\\
    multi-samp.$+$val. guidance                      & 0.538$\pm$0.010 & 0.542$\pm$0.031\\
    w/o PC to BEV projector                         &    N/A & 0.486$\pm$0.036\\
    \bottomrule
  \end{tabular}}
  \caption{Ablation experiments on navigation success rate in 15$\times$15 GridMaze2D and Home\_001\_1 scene of AVD.}
  \label{table:tab_ablation}
\end{table}
we replace the voting mechanism for plan selection with a learned value function. Instead of choosing the most frequently sampled plan, we select the one with the highest multi-step $Q$ value, as determined by the value function. This change markedly enhances performance, elevating success rates from 0.114 and 0.082 to 0.538 and 0.542 for GridMaze2D and AVD, respectively. This highlights the essential role of value-based guidance in our model's effectiveness.

\noindent\textbf{Effect of best plan memory.} We explore the significance of backtracking the past best plan, based on multi-sampling and value guidance. This mechanism is responsible for the performance gap between multi-samp.$+$val. guidance and the full version. The best plan memory addresses the issue of an optimal plan being replaced by a suboptimal one during re-planning in the context of receding horizon control. This underscores its crucial role in our methodology.

\noindent\textbf{Effect of point cloud to BEV projector.} Eliminating the semantic-segmentation-based projector hinders our framework's ability to apply the pre-trained policy for 2D domain to 3D navigation, necessitating the development of a new 3D-specific policy. To maintain the backbone of the diffusion-based plan generator, we adopt the lattice point net (LPN) used in CALVIN-3D for end-to-end policy learning. The complexity of this network alteration, coupled with the lack of ground-truth 2D maps for supervised projector training, leads to training difficulties, which causes a drop in success rate from 0.776 to 0.486. This emphasizes the importance of the semantic-segmentation-based projector in enabling the 2D policy's zero-shot transfer to 3D navigation.
\section{Conclusion}
\label{sec:conclusion}
This paper introduces a novel value-guided diffusion approach for trajectory-level plan generation, adept at navigating complex, long-horizon challenges under partial observability. Our approach exhibits remarkable versatility in both 2D and 3D environments and outperforms state-of-the-art methods. Extensive ablations underscore the importance of key constituents. Notably, our method effectively addresses the uncertainties inherent in partially observable environments, which is promising for real-world applications.

\noindent \textbf{Acknowledgments:} The first author especially thanks Bin Duan for his support during both submission and rebuttal. This research is supported by NSF IIS-2309073 and ECCS-212352101. This article solely reflects the opinions and conclusions of its authors and not the funding agencies.

\clearpage
{
    \small
    \bibliographystyle{ieeenat_fullname}
    \bibliography{main}
}

\clearpage
\setcounter{page}{1}
\maketitlesupplementary

\begin{table*}[t]
    \small
    \centering
    \begin{tabular}{lrrr}
        \toprule
        &\multicolumn{1}{c}{Down} & \multicolumn{1}{c}{Middle} & \multicolumn{1}{c}{Up} \\
        \midrule
        Input & 8$\times$4 & 2$\times$2048 & 2$\times$4096 \\
        \midrule
        \multirow{3}{*}{Stage 1} & Conv1d: [3, 1, 256]$\times$2 & Conv1d: [3, 1, 2048]$\times$2 & Conv1d: [3, 1, 1024]$\times$2 \\
        & Res1d: [3, 1, 256]$\times$1 & Res1d: [3, 1, 2048]$\times$1 & Res1d: [3, 1, 1024]$\times$1 \\
        & Downsample: [3, 2, 1024]$\times$1 & & Upsample: [4, 2, 1024]$\times$1 \\
        \midrule
        \multirow{3}{*}{Stage 2} & Conv1d: [3, 1, 1024]$\times$2 & Conv1d: [3, 1, 2048]$\times$2 & Conv1d: [3, 1, 256]$\times$2 \\
        & Res1d: [3, 1, 1024]$\times$1 & Res1d: [3, 1, 2048]$\times$1 & Res1d: [3, 1, 256]$\times$1 \\
        & Downsample: [3, 2, 2048]$\times$1 & & Upsample: [4, 2, 256]$\times$1 \\
        \midrule
        \multirow{2}{*}{Stage 3} & Conv1d: [3, 1, 2048]$\times$2 & & Conv1d: [3, 1, 256]$\times$1 \\
        & Res1d: [3, 1, 2048]$\times$1 & & Conv1d: [1, 1, 4]$\times$1 \\
        \midrule
        Output & 2$\times$2048 & 2$\times$2048 & 8$\times$4 \\
        \bottomrule
    \end{tabular}
    \caption{U-Net constitution of the plan generation module. Suppose the input action sequence length is 8 and has 4 channels. 4 is the number of bits used to encode each action. All our network's foundational components except upsampling are constructed using one-dimensional convolutional layers. Upsampling utilizes one-dimensional transposed convolutional layers. Within a convolutional layer, we designate the kernel size $K$, stride $S$, and target channels $C$ in the form of $[K,S,C]$. Additionally, we specify the input and output channels corresponding to each convolutional layer, which denotes the number of filters used. Note that the final convolutional block is regular instead of residual.}
    \label{table:unet-policy}
\end{table*}

\section{Network Architectural and Experimental Specification}
\label{sec:net_arch_exp}
\noindent\textbf{Expert trajectory padding.} To ensure alignment among different trajectories within expert demonstrations for closed-loop plan generation, we employ a padding strategy. Specifically, for the action trajectory, $\bm{\tau}_a$, we append $T_h$ steps of the termination action, \emph{Done}, to its end, aligning with the prediction horizon defined in \cref{subsec:method:diffusion_policy}. This trajectory is conditioned on the partial environmental map $\bm{e}$, corresponding to the timestep of the sequence's initial action. Thus, we pad at the end to match the prediction horizon's length. This approach, also functioning as a form of data augmentation, effectively addresses the sparsity of the \emph{Done} action in expert demonstrations, essential for training the policy to execute \emph{Done} accurately and timely. These adjustments ensure consistent alignment across $\bm{\tau}_{\bm{e}}$ and $\bm{\tau}_a$ in training data, with the same approach adopted during evaluation to condition the diffusion models properly.

\noindent\textbf{Partial maps.} Each partial map comprises three channels. The first marks observed obstacles as $1$s while marking free spaces and unobserved areas as $0$s. The second inversely marks observed free spaces as $1$s and the others as $0$s. The third marks the goal as $1$ if having observed it; otherwise, the channel is all-zero. We follow Wilson's methodology \cite{wilson1996randomspanningtree} to generate these maps. The starting point and target are selected uniformly from free cells, always ensuring a clear path between them. Under partial observability, the initial map presents minimal environmental information akin to the agent's first observation. As the exploration proceeds, the agent uncovers more areas and incorporates them into the progressively more complete map.

\noindent\textbf{Bit encoding.} We can represent discrete data from an alphabet of size $K$ variables with $n=\lceil\log_2K\rceil$ bits, in the form of $\{0, 1\}^n$. This discretization drives previous work to remodel continuous diffusion models to accommodate discrete data and state spaces. Nonetheless, Chen et al. put forward an alternative strategy, where binary bits, $\{0, 1\}^n$, are converted to real numbers, $\mathbb{R}^n$, thus making them suitable for continuous diffusion models. They refer to these real numbers as ``analog bits'', which are processed as real numbers despite mimicking the dual nature of binary bits. For sample generation, the approach remains the same as in continuous diffusion models but incorporates an additional step of quantization at the end, through which the resultant analog bits are thresholded. Consequently, this transforms the analog bits back into binary bits, ready to be turned into their original discrete or categorical variables.

\noindent\textbf{Plan generation module.} Our plan generation module employs a diffusion model anchored on a U-Net architecture. Each section of the U-Net, including the downward, middle, and upward modules, contains residual blocks merged with one-dimensional convolution neural networks (CNNs). \cref{table:unet-policy} showcases the specific layout of the plan generation module. The module takes in an expert action trajectory, with channels equal to the number of bits used to encode the action sequence (set to 4 for both domains) and length equal to $T_h$, serving as the input for the initial residual block. The hidden layers progressively widen the feature map channels to 256, 1024, and 2048 along the downsampling path. This map then navigates a bottleneck stage without altering its shape. In the upsampling path, the feature map is first concatenated with its counterpart stored during the downsampling path. Consequently, the input dimension for each stage of the upward module is effectively doubled. The final convolutional block of the upward module restores the output from a 256-channel feature map to a valid action trajectory. Note that no downsampling or upsampling is involved in the last stage of both paths.

An integral part of the plan generation module is an environment encoder, which converts a partial environment map at the current timestep, $\bm{e}_{(t)}$, into a low-dimensional embedding. This encoder comprises three convolutional blocks, succeeded by a global average pooling layer and two fully-connected layers. Each convolutional layer is configured with a kernel size of 3, a stride and padding size of 1, and output channels set to 128, 256, and 256, respectively. We apply group normalization, with eight groups, when conducting the convolutions. The global average pooling layer flattens the result into a one-dimensional vector of length 256. The two fully connected layers, with a dimension of 256$\times$256, encode the features extracted from the map into an embedding as the condition for conditional diffusion. This embedding is subsequently incorporated into the one-dimensional diffusion model.

We leverage Feature-wise Linear Modulation (FiLM) to model the conditional distribution $p(\bm{\tau}_{a,(t)}|\bm{e}_{(t)})$. To this end, we set the output channels as twice the output of the residual block. This configuration allows us to treat half of the conditional embedding as scale and the other half as bias, facilitating a linear operation on the output of the first convolutional block within the residual block and then going through the rest. Adopting FiLM enables the diffusion policy to adjust its behavior based on specific features of the input partial environmental map, facilitating more effective learning of the action trajectory and consequently boosting its performance.

\begin{table}[tb]
    \small
    \centering
    \setlength{\tabcolsep}{2.2pt}
    \centering
    \begin{tabular}{lr}
        \toprule
        Hyperparameter & Value \\
        \midrule
        Dataset (vary with env size) \\
        \hspace{1em}Minimal episode length (15$\times$15) & 16 \\
        \hspace{1em}Maxmal path length (15$\times$15) & 50 \\
        Diffusion model \\
        \hspace{1em}Observation horizon ($T_o$) & 2 \\
        \hspace{1em}Prediction horizon ($T_h$) & 4 \\
        \hspace{1em}Execution horizon ($T_a$) & 2 \\
        \hspace{1em}Timestep embedding dimension & 256 \\
        \hspace{1em}Diffusion steps & 32 \\
        \hspace{1em}EMA decay & 0.995 \\
        \hspace{1em}Number of plan candidates & 24 \\
        Value function \\
        \hspace{1em}Value iteration rounds ($K$) & 60 \\
        \hspace{1em}Discount factor ($\gamma$) & 0.99 \\
        Training \\
        \hspace{1em}Batch size & 32 \\
        \hspace{1em}Gradient accumulation steps & 2 \\
        \hspace{1em}Initial learning rate (diffusion policy) & 2e-4 \\
        \hspace{1em}Initial Learning rate (value function) & 0.005 \\
        \hspace{1em}Optimizer & RAdam \\
        \hspace{1em}Total training epochs & 60 \\
        \hspace{1em}Training steps per epoch & 10000 \\
        \bottomrule
    \end{tabular}
    \caption{Hyperparameters for training the diffusion-based planner and the value function estimator, respectively.}
    \label{table:hyperparams}
\end{table}

\noindent\textbf{Value function.} Within the value function module detailed in \cref{subsec:method:value_guidance}, the motion transition function $\hat{T}_m$, valid action reward function $\hat{R}_m$, and invalid action reward function $\hat{R}_f$ each consist of a convolutional kernel with $A$ filters. Notably, $\hat{T}_m$ is initialized as a Gaussian distribution with a Softmax function applied over the height and width, i.e., the action space $S$, to ensure a valid probability distribution. To enhance learning flexibility and representation, $\hat{T}_m$ has independent weights for belief update and value iteration. Both $\hat{R}_m$ and $\hat{R}_f$ are initialized to zero, and dot products in \cref{eq:pomdp-belief-update} and \cref{eq:QMDP_bellman} are implemented as convolutions using these kernels. The action embeddings $\hat{A}_{\rm{logit}}$ and $\hat{A}_{\rm{thresh}}$ are concurrently learned through CNNs. Specifically, the output action embedding comprises $(|A|+1)$ channels, with the first $|A|$ channels serving as $\hat{A}_{\rm{logit}}$ and the last one acting as $\hat{A}_{\rm{thresh}}$. We train the model to minimize the discrepancy between $\hat{A}_{\rm{logit}}$ and the expert demonstration, $a^{\ast}$, to reliably estimate the logarithmic probabilities associated with each action being taken.

\noindent\textbf{Parallel plan candidate generation.} One of our goals in adopting trajectory-level behavior synthesis for multi-step decision-making is to improve planning efficiency. However, introducing sampling variability, where the diffusion policy produces multiple plans for selection based on value function, ironically reintroduces inefficiency. This is because, at each planning step, the model must iteratively generate various plans. To address this, we employ multi-processing for plan generation, aligning the number of plans with the count of CPU cores. In our experiments, this number is set to 24 (refer to \cref{table:hyperparams}), effectively matching the time needed for generating multiple plans in a multi-processing setup to producing a single plan in a single-processing framework. This approach successfully preserves the high efficiency we initially sought.

\noindent\textbf{Hyperparameters.} We divide all the hyperparameters listed in \cref{table:hyperparams} into four parts, each corresponding to expert demonstration and dataset, diffusion policy module, value function module, and general training configurations.

\noindent\textbf{Transforming FPV RGB-D observations into 3D point clouds.} Our framework for 3D navigation first processes each depth image by creating a mesh grid based on the camera's intrinsic parameters, representing the camera's 2D image space pixel coordinates. Integrating this mesh grid with depths facilitates a 2D-to-3D mapping for each pixel. By employing the depth values and the camera mesh grid, the system calculates the 3D position of each pixel relative to the camera, a process known as unprojection. This method converts 2D pixel coordinates into 3D coordinates relative to the camera. Subsequently, these 3D coordinates are transformed into world coordinates using rotation and translation matrices, adjusting for the camera's position and orientation within the scene. This conversion process is systematically applied to each specified image. For every image processed, a 3D point cloud in world coordinates is generated and aggregated into a unified data structure. The culmination of this process is an extensive point cloud that encapsulates the entire scene as captured by the FPV images, where each point in the cloud corresponds to a pixel in the original depth images and is positioned according to its real-world location.

\noindent\textbf{Generating BEV environmental maps from 3D point cloud.} In \cref{subsec:method:3d_to_2d_projection}, we detail our use of Swin3D for semantic segmentation of 3D point clouds. The pre-trained Swin3D classifies points in unseen scenes into 13 categories: ceiling, floor, wall, beam, column, window, door, table, chair, sofa, bookcase, board, and clutter. Points unassigned to the first 12 categories default to clutter. After segmenting the point cloud (represented as $(X,Y,Z,R,G,B)$) with Swin3D, each point's class prediction is appended to its original tuple, creating a seven-element tuple. These tuples are then projected onto the X-Z plane, aligning with the BEV perspective, as the Y-axis (height in FPV RGB-D input) is perpendicular to the BEV plane. The Z-axis, indicating depth in FPV images, corresponds to one of the BEV dimensions. In our experiment, we map these points onto a grid determined by a given map resolution of 50$\times$50. For each grid cell, we classify it as an obstacle or free space based on the majority of points it contains—cells primarily with ceiling and floor points are marked as free space; others are deemed obstacles. This process yields a binary BEV map analogous to those in GridMaze2D, directly applying to our diffusion policy for planning.

\section{Additional Qualitative Results}
\begin{figure}[!h]\small 
    \centering
    \includegraphics[width=0.9\linewidth]{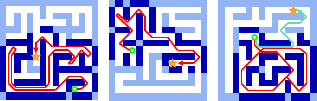}
    \caption{\textbf{Qualitative results in GridMaze2D.} The first two subfigures depict successful cases that use our approach. The last one illustrates a failed case, where the agent gets stuck in an indefinite loop. Refer to \cref{fig:experiment_example} for the denotation of markings.}
    \label{fig:qualitative}
\end{figure}

In \cref{fig:qualitative}, we present three cases of employing our method for navigation in GridMaze2D. The first two subfigures showcase successful navigation, where the agent following our policy explores the environment effectively, safely backtracks from dead ends, and finally reaches the target. Such cases account for the majority of all test cases. The third subfigure illustrates a failed scenario where the agent ends up in an indefinite loop. Unlike CALVIN's failure, where the agent repeats reverse actions consecutively and thus gets stuck in some small space, our policy leads to larger navigation path cycles, which are more unlikely to occur.

\vspace{-0.1cm}
\section{Additional Quantitative Results}

We evaluate the robustness against sensor noise in AVD. Specifically, we add noise sampled from $\mathcal{N}(0,\mathbf{I})$ to depth images. Numerical results in \cref{table:tab_avd_res} reveal a nuanced landscape of performance resilience. Both CALVIN variants exhibit a moderate decline in effectiveness under noisy conditions, with CALVIN-3D displaying slightly better resilience than its 2D counterpart, suggesting that 3D models possess inherent robustness to sensor inaccuracies in its trained 3D-to-2D projector. The zero-shot version of our approach stands out for its remarkable stability, experiencing negligible performance degradation despite the introduction of noise, underscoring its exceptional capability for zero-shot policy transfer in adverse environments. However, our policy retrained with RGB-D inputs consistently outperforms the others across a spectrum of scenarios and demonstrates extraordinary robustness. It suffers only minimal setbacks in the face of sensor noise. Thus, while all methods show some susceptibility to sensor noise, our approach emerges as the most robust, underlining the efficacy of incorporating additional sensory data through retraining in enhancing noise immunity. This finding could be pivotal for real-world applications where conditions are far from ideal.

\begin{table}[!h]
    \small
    \centering
    \setlength{\tabcolsep}{2.0pt}
    \scalebox{0.7}{\begin{tabular}{lcccc}
    \toprule
    Scene                   & CALVIN-2D & CALVIN-3D   & Ours (Zero-shot)    & Ours (Retrain)\\
    \midrule
    Home\_001\_1         & 0.692$\pm$0.037      & 0.720$\pm$0.052  &        0.769$\pm$0.038   & \textbf{0.776$\pm$0.028}\\
    Home\_001\_1 (noisy)  &  0.674$\pm$0.048  &  0.714$\pm$0.059  &  0.767$\pm$0.043    &  \textbf{0.772$\pm$0.033} \\
    Home\_001\_2         & 0.627$\pm$0.037      & 0.640$\pm$0.048  &  0.655$\pm$0.033   & \textbf{0.732$\pm$0.030} \\
    Home\_001\_2 (noisy)  &  0.604$\pm$0.042  &  0.638$\pm$0.058  &  0.652$\pm$0.036  &    \textbf{0.730$\pm$0.031} \\
    Home\_002\_1         & 0.735$\pm$0.035      & 0.740$\pm$0.048  &  0.728$\pm$0.034   & \textbf{0.755$\pm$0.027} \\
    Home\_002\_1 (noisy)  &  0.696$\pm$0.045  &  0.735$\pm$0.050  &  0.727$\pm$0.033  &   \textbf{0.754$\pm$0.030} \\
    Home\_003\_1         & 0.606$\pm$0.042      & 0.642$\pm$0.060  &  0.638$\pm$0.041   & \textbf{0.686$\pm$0.031} \\
    Home\_003\_1 (noisy)  &  0.613$\pm$0.050  &  0.636$\pm$0.059  &  0.634$\pm$0.043  &    \textbf{0.681$\pm$0.038} \\
    Home\_003\_2         & 0.558$\pm$0.033      & 0.590$\pm$0.043  &  0.603$\pm$0.033   &          \textbf{0.622$\pm$0.030} \\
    Home\_003\_2 (noisy)  &  0.552$\pm$0.036  &  0.589$\pm$0.048  &  0.602$\pm$0.034  &   \textbf{0.620$\pm$0.035} \\
    Home\_004\_1         & 0.647$\pm$0.040      & 0.680$\pm$0.050  &  0.684$\pm$0.042   &          \textbf{0.695$\pm$0.036} \\
    Home\_004\_1 (noisy)  &  0.634$\pm$0.043  &  0.674$\pm$0.057  &  0.681$\pm$0.042  &   \textbf{0.690$\pm$0.043} \\
    Home\_007\_1         & 0.587$\pm$0.038      & \textbf{0.610$\pm$0.045}  & 0.584$\pm$0.039   & 0.601$\pm$0.035 \\
    Home\_007\_2 (noisy)  &  0.580$\pm$0.041  &  \textbf{0.602$\pm$0.049}  &  0.584$\pm$0.040  &    0.598$\pm$0.038 \\
    Home\_010\_1         & 0.728$\pm$0.033      & 0.736$\pm$0.043  &  0.769$\pm$0.032   & \textbf{0.781$\pm$0.028} \\
    Home\_010\_1 (noisy)  &  0.717$\pm$0.030  &  0.731$\pm$0.046  &  0.766$\pm$0.033  &    \textbf{0.780$\pm$0.032} \\
    Mean succ. rate      & 0.635$\pm$0.032 & 0.682$\pm$0.047  & 0.679$\pm$0.040 & \textbf{0.706$\pm$0.032} \\
    Mean succ. rate (noisy)  &  0.626$\pm$0.037  &  0.670$\pm$0.052   &  0.675$\pm$0.042  &  \textbf{0.700$\pm$0.040} \\
    \bottomrule
    \end{tabular}}
    \caption{Performance of CALVIN and our method in AVD's embodied navigation and object searching tasks, where the goal is to locate a Coca-Cola glass bottle in an indoor scene. It presents the agent's success rates across various scenes. Our method, which achieves comparable results to CALVIN in zero-shot policy transfer from the 2D domain, surpasses CALVIN in scenarios retrained with additional RGB inputs, with an exception in one scene.}
    \label{table:tab_avd_res}
\end{table}

\section{Limitations and Future Work}
This work has two main limitations—high reliance on precise point cloud semantic segmentation for the 3D domain and potential suboptimal long-term decision-making embedded in the value guidance due to QMDP's strong assumption of full observability in future timesteps. 

\noindent\textbf{Reliance on the performance of segmentation model.} For effective semantic segmentation of point clouds, it is imperative that the chosen model reliably identifies key objects like floors, ceilings, and walls across diverse indoor settings. Mislabeling can lead to incorrect BEV map projections and cause catastrophic planning errors. Continuously improving the segmentation model's quality is essential to ensure accuracy. Thus, future work may involve fine-tuning the model on a small, labeled dataset from the target domain, if available, to boost performance significantly. Alternatively, for domains where labeled data is unavailable, unsupervised adaptation methods like adversarial training or self-ensembling could be employed. However, developing an entirely new semantic segmentation framework specifically for point cloud data is beyond the scope of our discussion, which focuses on decision-making in navigation planning.

\noindent\textbf{Potential suboptimal long-term planning.} As we introduce in \cref{sec:related_work}, the QMDP heuristic simplifies the value iteration of POMDP by taking into account partial observability only at the current timestep but assuming full observability on subsequent timesteps. Doing so can make optimal decisions in the immediate sense but be suboptimal when considering a longer horizon. In our work, unlike autoregressive approaches, we need not only one optimal step of action but a sequence of steps that accounts for the predicted action trajectory, which might amplify the impact of suboptimality of QMDP in farther timesteps. To address this issue, we can consider point-wise value iteration in future work.

\clearpage

\end{document}